\documentclass[runningheads]{llncs}
\usepackage{graphicx}
\usepackage{tikz}
\usepackage{amsmath,amssymb} 
\usepackage{color}

\usepackage{wasysym}
\usepackage{adjustbox}
\usepackage{multirow}
\usepackage{wrapfig}
\usepackage{subfig}
\usepackage{csquotes}
\usepackage[pagebackref=true,breaklinks=true,colorlinks,bookmarks=false,citecolor=blue,linkcolor=blue]{hyperref}

\def\eg{\emph{e.g}.} 
\def\Eg{\emph{E.g}.}

\newcommand{\clswgan}{\texttt{f-CLSWGAN}}
\newcommand{\vaegan}{\texttt{f-VAEGAN}}
\newcommand{\featcat}{\texttt{T-feature}}
\newcommand{\feedback}{\texttt{Feedback}}
\newcommand{\proposed}{\texttt{TF-VAEGAN}}
\newcommand\blfootnote[1]{%
  \begingroup
  \renewcommand\thefootnote{}\footnote{#1}%
  \addtocounter{footnote}{-1}%
  \endgroup
}

\begin{document}

\pagestyle{headings}
\mainmatter
\def\ECCVSubNumber{}  

\title{Latent Embedding Feedback and Discriminative Features for Zero-Shot Classification} 

\titlerunning{Latent Embedding Feedback and Discriminative Features for ZSC}

\authorrunning{S. Narayan, A. Gupta, F. S. Khan, C. G. M. Snoek, L. Shao}

\author{Sanath Narayan*$^1$, Akshita Gupta*$^1$, Fahad Shahbaz Khan$^{1,3}$, \\
Cees G. M. Snoek$^2$, Ling Shao$^{1,3}$}
\institute{
  $^1$~Inception Institute of Artificial Intelligence, UAE \quad
  $^2$~University of Amsterdam
  $^3$~Mohamed Bin Zayed University of Artificial Intelligence, UAE
  }

\maketitle

\begin{abstract}
\blfootnote{$*$ Equal Contribution. Correspondence: sanath.narayan@inceptioniai.org}Zero-shot learning strives to classify unseen categories for which no data is available during training. In the generalized variant, the test samples can further belong to seen or unseen categories. The state-of-the-art relies on Generative Adversarial Networks that synthesize unseen class features by leveraging class-specific semantic embeddings. During training, they generate semantically consistent features, but discard this constraint during feature synthesis and classification. We propose to enforce semantic consistency at \emph{all} stages of (generalized) zero-shot learning: training, feature synthesis and classification. We first introduce a feedback loop, from a semantic embedding decoder, that iteratively refines the generated features during both the training and feature synthesis stages. 
The synthesized features together with their corresponding latent embeddings from the decoder are then transformed into discriminative features and utilized during classification to reduce ambiguities among categories. 
Experiments on (generalized) zero-shot object and action classification reveal the benefit of semantic consistency and iterative feedback, outperforming existing methods on six zero-shot learning benchmarks. 
Source code at \url{https://github.com/akshitac8/tfvaegan}.
\end{abstract}

\section{Introduction}

This paper strives for zero-shot learning, a challenging vision problem that involves classifying images or videos into new (\enquote{unseen}) categories at test time, without having been provided any corresponding visual example during training. In the literature~\cite{akata15label,romera15icml,Ye17cvpr,Xian18cvpr}, this is typically achieved by utilizing the labelled seen class instances and class-specific semantic embeddings (provided as a side information), which encode the inter-class relationships. Different from the zero-shot setting, the test samples can belong to the seen or unseen categories in generalized zero-shot learning~\cite{zsl-good-bad-ugly}. In this work, we investigate the problem of both zero-shot learning (ZSL) and generalized zero-shot learning (GZSL).

Most recent ZSL and GZSL recognition approaches~\cite{Xian18cvpr,Rafael18eccv,Xian19cvpr,huang19generative,li19leveraging} are based on Generative Adversarial Networks (GANs)~\cite{GAN}, which aim at directly optimizing the divergence between real and generated data. The work of~\cite{Xian18cvpr} learns a GAN using the seen class feature instances and the corresponding class-specific semantic embeddings, which are either manually annotated or word vector~\cite{mikolov13w2v} representations. 
Feature instances of the unseen categories, whose real features are unavailable during training, are then synthesized using the trained GAN and used together with the real feature instances from the seen categories to train zero-shot classifiers in a fully-supervised setting. 
A few works~\cite{Rafael18eccv,huang19generative,Mandal19cvpr} additionally utilize auxiliary modules, such as a decoder, to enforce a cycle-consistency constraint on the reconstruction of semantic embeddings during training. Such an auxiliary decoder module aids the generator to synthesize semantically consistent features.
Surprisingly, these modules are \textit{only} employed during training and discarded during \textit{both} the feature synthesis and ZSL classification stages. Since the auxiliary module aids the generator during training, it is also expected to help obtain discriminative features during feature synthesis \textit{and} reduce the ambiguities among different classes during classification. In this work, we address the issues of enhanced feature synthesis and improved zero-shot classification.

Further, GANs are likely to encounter mode collapse issues~\cite{arjovsky2017towards}, resulting in decreased diversity of generated features. While Variational Autoencoders (VAEs)~\cite{kingma13iclr} achieve more stable feature generation, the approximate inference distribution is likely to be different from the true posterior~\cite{zhao2019infovae}. Recently, \cite{Xian19cvpr} build on~\cite{Xian18cvpr} to combine the strengths of VAEs and GANs and introduce an \vaegan{} ZSL framework by sharing the VAE decoder and GAN generator modules. To ensure that the generated features are semantically close to the distribution of real feature, a cycle-consistency loss~\cite{CycleGAN} is employed between generated and original features, during training. Here, we propose to additionally enforce a similar consistency loss on the semantic embeddings during training and further utilize the learned information during feature synthesis and classification.

\subsection{Contributions}
We propose a novel method, which advocates the effective utilization of a semantic embedding decoder (SED) module at \textit{all} stages of the ZSL framework: training, feature synthesis and classification. Our method is built on a VAE-GAN architecture. (i)~We design a \textit{feedback module} for (generalized) zero-shot learning that utilizes SED during both training and feature synthesis stages. The feedback module first transforms the latent embeddings of SED, which are then used to modulate the latent representations of the generator. To the best of our knowledge, we are the first to propose a feedback module, within a VAE-GAN architecture, for the problem of (generalized) zero-shot recognition. (ii) We introduce a \textit{discriminative feature transformation}, during the classification stage, that utilizes the latent embeddings of SED along with their corresponding visual features for reducing ambiguities among object categories. In addition to object recognition, we show effectiveness of the proposed approach for (generalized) zero-shot action recognition in videos.

We validate our approach by performing comprehensive experiments on four commonly used ZSL object recognition datasets: CUB~\cite{cub}, FLO~\cite{flo}, SUN~\cite{sun} and AWA~\cite{zsl-good-bad-ugly}. Our experimental evaluation shows the benefits of utilizing SED at all stages of the ZSL/GZSL pipeline. In comparison to the baseline, the proposed approach obtains absolute gains of 4.6\%, 7.1\%, 1.7\%, and 3.1\% on CUB, FLO, SUN, and AWA, respectively for generalized zero-shot (GZSL) object recognition. In addition to object recognition, we evaluate our method on two (generalized) zero-shot action recognition in videos datasets: HMDB51~\cite{hmdb51} and UCF101~\cite{ucf101}. Our approach outperforms existing methods on \textit{all} six datasets. We also show the generalizability of our proposed contributions by integrating them into GAN-based (generalized) zero-shot recognition framework.

\section{Related Work}

In recent years, the problem of object recognition under zero-shot learning (ZSL) settings has been well studied~\cite{Jayaraman14nips,fu15pami,akata15label,frome13nips,romera15icml,rohrbach13nips,Ye17cvpr,Xian18cvpr}. Earlier ZSL image classification works~\cite{Jayaraman14nips,lampert13pami} learn semantic embedding classifiers for associating seen and unseen classes. Different from these methods, the works of~\cite{akata15label,frome13nips,romera15icml} learn a compatibility function between the semantic embedding and visual feature spaces. Other than these inductive approaches that rely only on the labelled data from seen classes, the works of~\cite{fu15pami,rohrbach13nips,Ye17cvpr} leverage additional unlabelled data from unseen classes through label propagation under a transductive zero-shot setting.

Recently, Generative Adversarial Networks~\cite{GAN} (GANs) have been employed to synthesize unseen class features, which are then used in a fully supervised setting to train ZSL classifiers~\cite{Xian18cvpr,Rafael18eccv,li19leveraging,Xian19cvpr}. A conditional Wasserstein GAN~\cite{wgan} (WGAN) is used along with a seen category classifier to learn the generator for unseen class feature synthesis~\cite{Xian18cvpr}.~This is achieved by using a WGAN loss and a classification loss.~In~\cite{Rafael18eccv}, the seen category classifier is replaced by a decoder together with the integration of a cycle-consistency loss~\cite{CycleGAN}.~The work of~\cite{schonfeld19cvpr} proposes an approach where cross and distribution alignment losses are introduced for aligning the visual features and corresponding embeddings in a shared latent space, using two Variational Autoencoders~\cite{kingma13iclr} (VAEs).~The work of~\cite{Xian19cvpr} introduces a \vaegan~framework which combines a VAE and a GAN by sharing the decoder of VAE and generator of GAN for feature  synthesis.  For training, the \vaegan~framework utilizes a cycle-consistency constraint between generated and original visual features. However, a similar constraint is not enforced on the semantic embeddings in their framework. Different from \vaegan, other GAN-based ZSL classification methods~\cite{Rafael18eccv,Zhang18ijcai,huang19generative,Mandal19cvpr} investigate the utilization of auxiliary modules to enforce cycle-consistency on the embeddings. Nevertheless, these modules are utilized only during training and discarded during both feature synthesis and ZSL classification stages.

Previous works~\cite{amir17cvpr,huh19feedbackcvpr,li19feedbackcvpr,shama19iccv} have investigated leveraging feedback information to incrementally improve the performance of different applications, including classification, image-to-image translation and super-resolution. To the best of our knowledge, our approach is the first to incorporate a feedback loop for improved feature synthesis in the context of (generalized) zero-shot recognition (both image and video). We systematically design a feedback module, in a VAE-GAN framework, that iteratively refines the synthesized features for ZSL.

While zero-shot image classification has been extensively studied, zero-shot action recognition in videos received less attention. Several works~\cite{Kodirov15iccv,Xu17ijcv,Mishra18wacv} study the problem of zero-shot action recognition in videos under transductive setting. The use of image classifiers and object detectors for action recognition under ZSL setting are investigated in~\cite{jain15iccv,mettes17iccv}. 
Recently, GANs have been utilized to synthesize unseen class video features in~\cite{Zhang18ijcai,Mandal19cvpr}. 
Here, we further investigate the effectiveness of our framework for zero-shot action recognition in videos.

\section{Method}

We present an approach, \proposed, for (generalized) zero-shot recognition. As discussed earlier, the objective in ZSL is to classify images or videos into new classes, which are unknown during the training stage. Different from ZSL, test samples can belong to seen or unseen classes in the GZSL setting, thereby making it a harder problem due to the domain shift between the seen and unseen classes. 
Let $x \in \mathcal{X}$ denote the encoded feature instances of images (videos) and $y \in \mathcal{Y}^s$ the corresponding labels from the set of $M$ seen class labels $\mathcal{Y}^s = \{y_1,\ldots,y_M\}$. Let $\mathcal{Y}^u = \{u_1,\ldots,u_N\}$ denote the set of $N$ unseen classes, which is disjoint from the seen class set $\mathcal{Y}^s$. The seen and unseen classes are described by the category-specific semantic embeddings $a(k) \in \mathcal{A}$, $\forall k \in \mathcal{Y}^s \cup \mathcal{Y}^u$, which encode the relationships among all the classes. 
While the unlabelled test features $x_t \in \mathcal{X}$ are not used during training in the inductive setting, they are used during training in the transductive setting to reduce the bias towards seen classes.
The tasks in ZSL and GZSL are to learn the classifiers $f_{zsl}: \mathcal{X} \rightarrow \mathcal{Y}^u $ and $f_{gzsl}: \mathcal{X} \rightarrow \mathcal{Y}^s \cup \mathcal{Y}^u $, respectively. To this end, we first learn to synthesize the features using the seen class features $x_s$ and corresponding embeddings $a(y)$. The learned model is then used to synthesize unseen class features $\hat{x}_u$ using the unseen class embeddings $a(u)$. The resulting synthesized features $\hat{x}_u$, along with the real seen class features $x_s$, are further deployed to train the final classifiers $f_{zsl}$ and $f_{gzsl}$.

\subsection{Preliminaries: \bf\vaegan\label{sec_vaegan}}
We base our approach on the recently introduced \vaegan~\cite{Xian19cvpr}, which combines the strengths of the VAE~\cite{kingma13iclr} and GAN~\cite{GAN} as discussed earlier, achieving impressive results for ZSL classification. Compared to GAN based models, \eg, \texttt{f-CLSWGAN}~\cite{Xian18cvpr}, the \vaegan~\cite{Xian19cvpr} generates semantically consistent features by sharing the decoder and generator of the VAE and GAN.
In \vaegan, the feature generating VAE~\cite{kingma13iclr} (\texttt{f-VAE}) comprises an encoder $E(x,a)$, which encodes an input feature $x$ to a latent code $z$, and a decoder $G(z,a)$ (shared with \texttt{f-WGAN}, as a conditional generator) that reconstructs $x$ from $z$. Both $E$ and $G$ are conditioned on the embedding $a$, optimizing, 
\begin{equation}
    \label{eqn_vae}
    \mathcal{L}_{V} = \mathrm{KL}(E(x,a)||p(z|a)) - \mathbb{E}_{E(x,a)}[\log G(z,a)],
\end{equation}
where $\mathrm{KL}$ is the Kullback-Leibler divergence, $p(z|a)$ is a prior distribution, assumed to be $\mathcal{N}(0,1)$ and $\log G(z,a)$ is the reconstruction loss. The feature generating network~\cite{Xian18cvpr} (\texttt{f-WGAN}) comprises a generator $G(z,a)$ and a discriminator $D(x,a)$. The generator $G(z,a)$ synthesizes a feature $\hat{x} \in \mathcal{X}$ from a random input noise $z$, whereas the discriminator $D(x,a)$ takes an input feature $x$ and outputs a real value indicating the degree of realness or fakeness of the input features. Both $G$ and $D$ are conditioned on the embedding $a$, optimizing the WGAN loss $\mathcal{L}_{W}= \mathbb{E}[D(x,a)] - \mathbb{E}[D(\hat{x},a)] - \lambda \mathbb{E}[(||\nabla D(\tilde{x},a)||_2 - 1)^2]$.
Here, $\hat{x} = G(z,a)$ is the synthesized feature, $\lambda$ is the penalty coefficient and $\tilde{x}$ is sampled randomly from the line connecting $x$ and $\hat{x}$. The \vaegan~is then optimized by:
\begin{equation}
    \label{eqn_vaegan}
    \mathcal{L}_{vaegan} = \mathcal{L}_V + \alpha \mathcal{L}_W,
\end{equation}
where $\alpha$ is a hyper-parameter. For more details, we refer to~\cite{Xian19cvpr}. 

\begin{figure}[t]
\centering
\includegraphics[width=0.9\columnwidth]{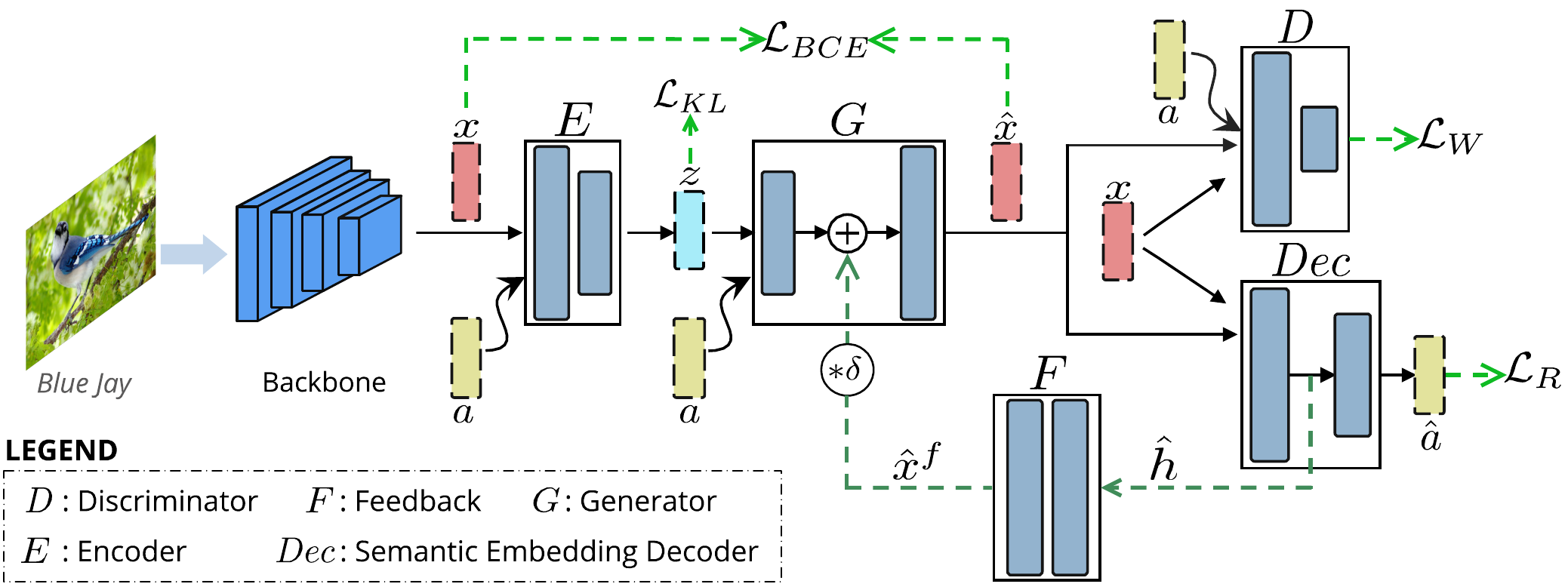}%
\caption{\label{fig_overall_arch}\textbf{Proposed architecture} (Sec~\ref{sec_overall_arch}). Given a seen class image, visual features $x$ are extracted from the backbone network and input to the encoder $E$, along with the corresponding semantic embeddings $a$. The encoder $E$ outputs a latent code $z$, which is then input together with embeddings $a$ to the generator $G$ that synthesizes features $\hat{x}$. The discriminator $D$ learns to distinguish between real and synthesized features $x$ and $\hat{x}$, respectively. Both $E$ and $G$ together constitute the VAE, which is trained using a binary cross-entropy loss ($\mathcal{L}_{BCE}$) and the KL divergence ($\mathcal{L}_{KL}$). Similarly, both $G$ and $D$ form the GAN trained using the WGAN loss ($\mathcal{L}_{W}$).  A semantic embedding decoder $Dec$ is introduced (Sec.~\ref{sec_dec_emb}) to reconstruct the embeddings $\hat{a}$ using a cycle-consistency loss ($\mathcal{L}_R$). Further, a feedback module $F$ (Sec.~\ref{sec_feedback}) is integrated to transform the latent embedding $\hat{h}$ of $Dec$ and feed it back to $G$, which iteratively refines $\hat{x}$.}%
\end{figure}%

\noindent\textbf{Limitations}: The loss formulation for training \vaegan, contains a constraint (second term in Eq.~\ref{eqn_vae}) that ensures the generated visual features are cyclically-consistent, at train time, with the original visual features. However, a similar cycle-consistency constraint is not enforced on the semantic embeddings. Alternatively, other GAN-based ZSL methods~\cite{Rafael18eccv,Zhang18ijcai} utilize auxiliary modules (apart from the generator) for achieving cyclic-consistency on embeddings. However, these modules are employed \textit{only} during training and discarded at both feature synthesis and ZSL classification stages. In this work, we introduce a semantic embedding decoder (SED) that enforces cycle-consistency on semantic embeddings and utilize it at \textit{all} stages: training, feature synthesis and ZSL classification. We argue that the generator and SED contain complementary information with respect to feature instances, since the two modules perform inverse transformations in relation to each other. The generator module transforms the semantic embeddings to the feature instances whereas, SED transforms the feature instances to semantic embeddings.~Our approach focuses on the utilization of this complementary information for improving feature synthesis and reducing ambiguities among classes (\eg, fine-grained classes) during ZSL classification.

\subsection{Overall Architecture\label{sec_overall_arch}}
The overall architecture is illustrated in Fig.~\ref{fig_overall_arch}. The VAE-GAN consists of an encoder $E$, generator $G$ and discriminator $D$. The input to $E$ are the real features of seen classes $x$ and the semantic embeddings $a$ and the output of $E$ are the parameters of a noise distribution. These parameters are matched to those of a zero-mean unit-variance Gaussian prior distribution using the KL divergence ($\mathcal{L}_{KL}$). The noise $z$ and embeddings $a$ are input to $G$, which synthesizes the features $\hat{x}$. The synthesized features $\hat{x}$ and original features $x$ are compared using a binary cross-entropy loss $\mathcal{L}_{BCE}$. The discriminator $D$ takes either $x$ or $\hat{x}$ along with embeddings $a$ as input, and computes a real number that determines whether the input is real or fake. The WGAN loss $\mathcal{L}_{W}$ is applied at the output of $D$ to learn to distinguish between the real and fake features.

The focus of our design is the integration of an additional semantic embedding decoder (SED) $Dec$ at both the feature synthesis and ZSL/GZSL classification stages. Additionally, we introduce a feedback module $F$, which is utilized during training and feature synthesis, along with $Dec$. Both the semantic embedding decoder $Dec$ and feedback module $F$ collectively address the objectives of enhanced feature synthesis and reduced ambiguities among categories during classification. The $Dec$ takes either $x$ or $\hat{x}$ and reconstructs the embeddings $\hat{a}$. It is trained using a cycle-consistency loss $\mathcal{L}_R$. The learned $Dec$ is subsequently used in the ZSL/GZSL classifiers. The feedback module $F$ transforms the latent embedding of $Dec$ and feeds it back to the latent representation of generator $G$ in order to achieve improved feature synthesis. The SED $Dec$ and feedback module $F$ are described in detail in Sec.~\ref{sec_dec_emb} and~\ref{sec_feedback}.

\subsection{Semantic Embedding Decoder\label{sec_dec_emb}}
Here, we introduce a semantic embedding decoder $Dec:\mathcal{X} \rightarrow \mathcal{A}$, for reconstructing the semantic embeddings $a$ from the generated features $\hat{x}$. Enforcing a cycle-consistency on the reconstructed semantic embeddings ensures that the generated features are transformed to the same embeddings that generated them. As a result, semantically consistent features are obtained during feature synthesis. The cycle-consistency of the semantic embeddings is achieved using the $\ell_1$ reconstruction loss as follows:
\begin{equation}
    \label{eqn_decoder}
    \mathcal{L}_{R} = \mathbb{E}[||Dec(x) - a||_1] + \mathbb{E}[||Dec(\hat{x}) - a||_1].
\end{equation}
The loss formulation for training the proposed \proposed{} is then given by,
\begin{equation}
    \label{eqn_total_loss}
    \mathcal{L}_{total} = \mathcal{L}_{vaegan} + \beta \mathcal{L}_R,
\end{equation}
\noindent where $\beta$ is a hyper-parameter for weighting the decoder reconstruction error. 

As discussed earlier, existing GAN-based ZSL approaches~\cite{Rafael18eccv,Zhang18ijcai} employ a semantic embedding decoder (SED) \textit{only} during training and discard it during \textit{both} unseen class feature synthesis and ZSL classification stage. In our approach, SED is utilized at \textit{all} three stages of VAE-GAN based ZSL pipeline: training, feature synthesis and classification. Next, we describe importance of SED during classification and later investigate its role during feature synthesis  (Sec.~\ref{sec_feedback}).

\begin{figure}[t]
\begin{minipage}{0.5\textwidth}
    \subfloat[Discriminative Feature Transformation]
    {
        \includegraphics[width=0.99\columnwidth]{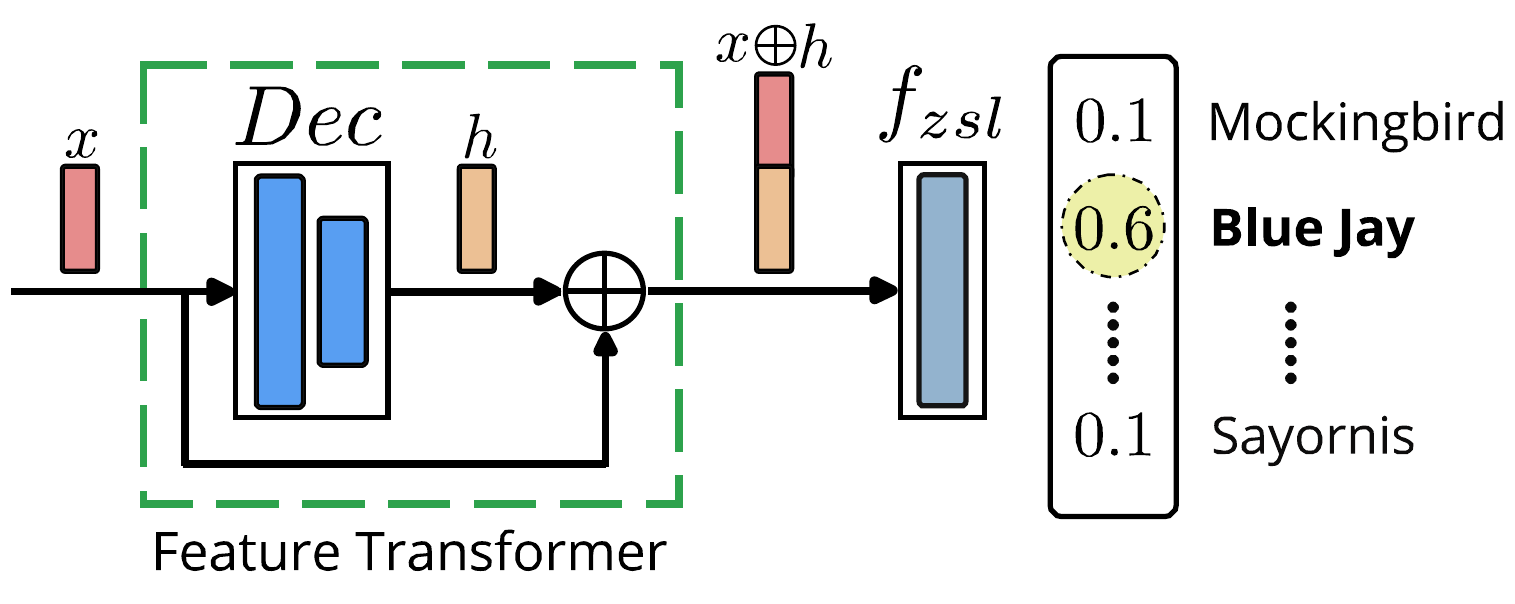}
        \label{fig_dec_cls}
    }
    \end{minipage}%
    \begin{minipage}{0.5\textwidth}
    \centering
    \subfloat[Feedback Module]
    {
        \includegraphics[width=0.99\columnwidth]{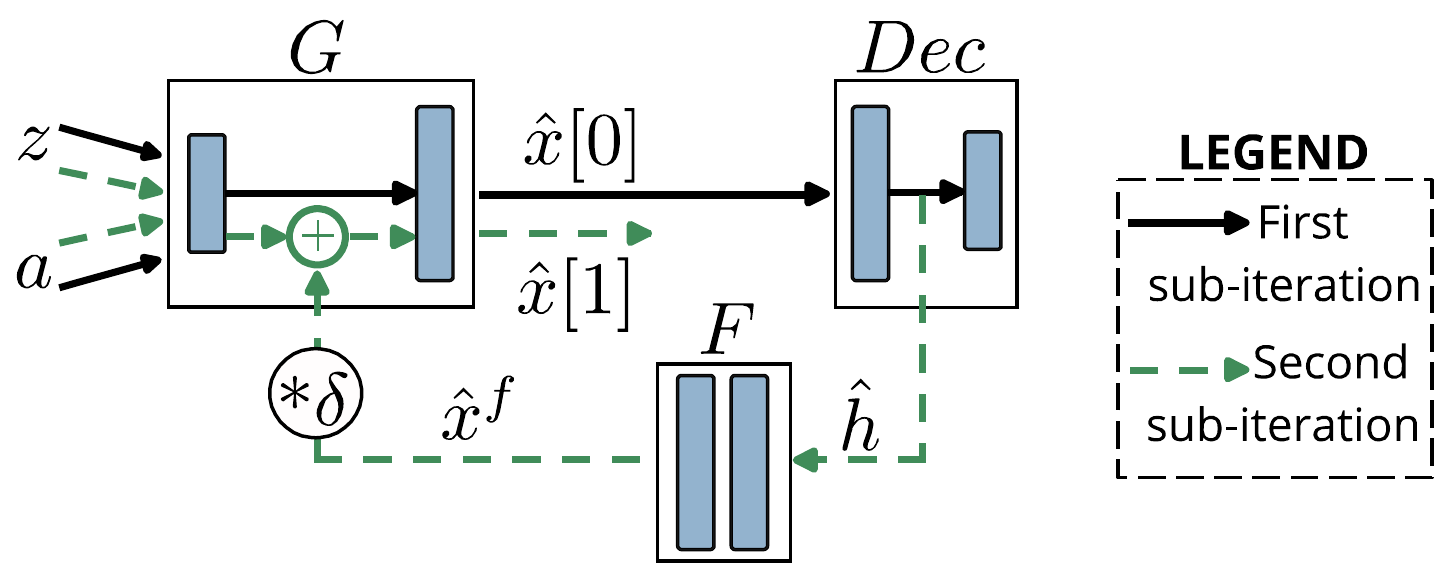}
        \label{feedback_fig}
    }
\end{minipage}%
\caption{(a) \textbf{Integration of semantic embedding decoder} $Dec$ at the ZSL/GZSL classification stage. A feature transformation is performed by concatenating ($\oplus$) the input visual features $x$ with the corresponding latent embedding $h$ from SED. The transformed discriminative features are then used for ZSL/GZSL classification. \\ (b) \textbf{Feedback module overview}. First sub-iteration: The generator $G$ synthesizes initial features $\hat{x}[0]$ using the noise $z$ and embeddings $a$. The initial features are passed through the $Dec$. Second sub-iteration: The module $F$ transforms the latent embedding $h$ from $Dec$ to $\hat{x}^f$, which represents the feedback to $G$. The generator $G$ synthesizes enhanced features $\hat{x}[1]$ using the same $z$ and $a$ along with the feedback $\hat{x}^f$.}%
\end{figure}

\noindent\textbf{Discriminative feature transformation:}
Here, we describe the proposed discriminative feature transformation scheme to effectively utilize the auxiliary information in semantic embedding decoder (SED) at the ZSL classification stage. The generator $G$ learns a \emph{per-class} \enquote{single semantic embedding to many instances} mapping using only the seen class features and embeddings.
Similar to the generator $G$, the SED is also trained using only the seen classes but learns a \emph{per-class} \enquote{many instances to one embedding} inverse mapping. Thus, the generator $G$ and SED $Dec$ are likely to encode complementary information of the categories.  Here, we propose to use the latent embedding from SED as a useful source of information at the classification stage (see Fig.~\ref{fig_dec_cls}) for reducing ambiguities among features instances of different categories. 

First, the training of feature generator $G$ and semantic embedding decoder $Dec$ is performed. Then, $Dec$ is used to transform the features (real and synthesized) to the embedding space $\mathcal{A}$. Afterwards, the latent embeddings from $Dec$ are concatenated with the respective visual features. Let $h_s$ and $\hat{h}_u \in \mathcal{H}$ denote the hidden layer (latent) embedding from the $Dec$ for inputs $x_s$ and $\hat{x}_u$, respectively. The transformed features are represented by: $x_s \oplus h_s$ and $\hat{x}_u \oplus \hat{h}_u$, where $\oplus$ denotes concatenation. In our method, the transformed features are used to learn final ZSL and GZSL classifiers as,
\begin{equation}
    \label{eqn_mod_classifiers}
    f_{zsl}:\mathcal{X} \oplus \mathcal{H}  \rightarrow \mathcal{Y}^u \qquad \text{and} \qquad f_{gzsl}:\mathcal{X} \oplus \mathcal{H} \rightarrow \mathcal{Y}^s \cup \mathcal{Y}^u.
\end{equation}
As a result, the final classifiers learn to better distinguish categories using transformed features.
Next, we describe integration of $Dec$ during feature synthesis.

\subsection{Feedback Module\label{sec_feedback}}
\begin{wrapfigure}{R}{0.5\columnwidth}
\centering
\includegraphics[width=0.5\columnwidth]{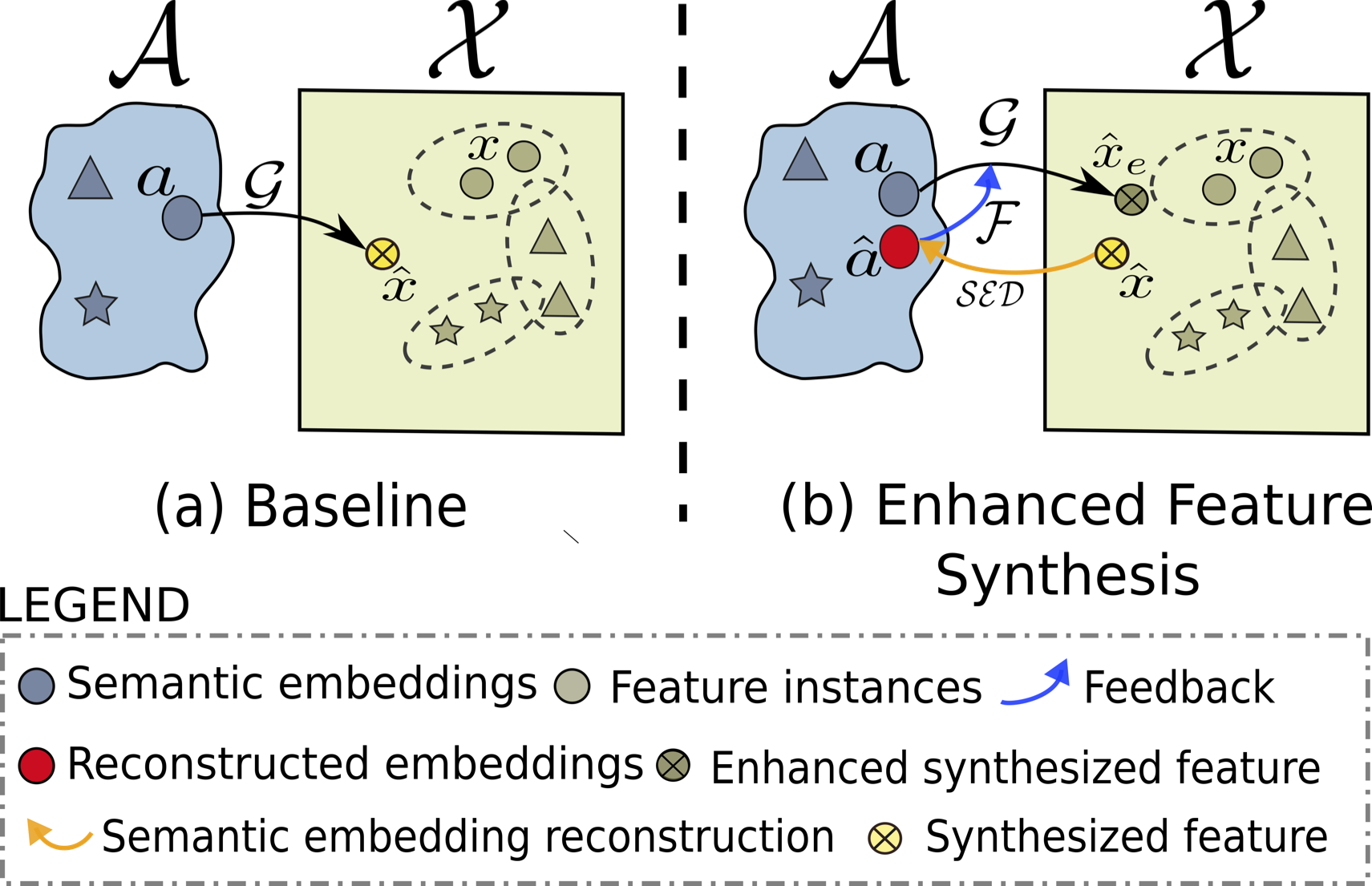}
\caption{\label{new_feedback_fig} \textbf{Conceptual illustration} between the baseline (a) and our feedback module designed for enhanced feature synthesis (b), using three classes ($\bigstar$, $\blacktriangle$ and $\CIRCLE$). The baseline learns to synthesize features $\hat{x}$ from class-specific semantic embeddings $a$ via generator $G$, without enforcing cycle-consistency in the attribute space. As a consequence, a semantic gap is likely to exist between the synthesized and real $x$ features. In our approach, cycle-consistency is enforced using SED. Further, the disparity between the reconstructed embeddings $\hat{a}$ and $a$ is used as a \textit{feedback signal} to reduce the semantic gap between $\hat{x}$ and $x$, resulting in enhanced synthesized features $\hat{x}_e$.}
\end{wrapfigure}

The baseline \vaegan~does not enforce cycle-consistency in the attribute space and directly synthesizes visual features $\hat{x}$ from the class-specific embeddings $a$ via the generator (see Fig.~\ref{new_feedback_fig}a). This results in a semantic gap between the real and synthesized visual features. To address this issue, we introduce a feedback loop that iteratively refines the feature generation (see Fig.~\ref{new_feedback_fig}b) during both the training and synthesis stages. The feedback loop is introduced from the semantic embedding decoder $Dec$ to the generator $G$, through our feedback module $F$ (see Fig.~\ref{fig_overall_arch} and Fig.~\ref{feedback_fig}). The proposed module $F$ enables the effective utilization of $Dec$ during both training and feature synthesis stages. Let $g^l$ denote the $l^{th}$ layer output of $G$ and $\hat{x}^f$ denote the feedback component that additively modulates $g^l$. The feedback modulation of output $g^l$ is given by,
\begin{equation}
    \label{eqn_feedback_concept}
    g^l \leftarrow g^l + \delta \hat{x}^f,
\end{equation}
where $\hat{x}^f = F(h)$, with $h$ as the latent embedding of $Dec$ and $\delta$ controls the feedback modulation. To the best of our knowledge, we are the first to design and incorporate a feedback loop for zero-shot recognition. Our feedback loop is based on~\cite{shama19iccv}, originally introduced for image super-resolution. However, we observe that it provides sub-optimal performance for zero-shot recognition due to its less reliable feedback during unseen class feature synthesis. Next, we describe an improved feedback loop with necessary modifications for zero-shot recognition. 

\noindent \textbf{Feedback module input}: The adversarial feedback employs a latent representation of an unconditional discriminator $D$ as its input~\cite{shama19iccv}.
However, in the ZSL problem, $D$ is conditional and is trained with an objective to distinguish between the real and fake features of the seen categories. This restricts $D$ from providing reliable feedback during unseen class feature synthesis. In order to overcome this limitation, we turn our attention to semantic embedding decoder $Dec$, whose aim is to reconstruct the class-specific semantic embeddings from features instances. Since $Dec$ learns class-specific transformations from visual features to the semantic embeddings, it is better suited (than $D$) to provide feedback to generator $G$. \\
\noindent \textbf{Training strategy}: Originally, the feedback module $F$ is trained in a two-stage fashion~\cite{shama19iccv}, where the generator $G$ and discriminator $D$ are first fully trained, as in the standard GAN training approach. Then, $F$ is trained using a feedback from $D$ and freezing $G$. Since, the output of $G$ improves due to the feedback from $F$, the discriminator $D$ is continued to be trained alongside $F$, in an adversarial manner. In this work, we argue that such a two-stage training strategy is sub-optimal for ZSL, since $G$ is always fixed and not allowed to improve its feature synthesis. To further utilize the feedback for improved feature synthesis, $G$ and $F$ are trained alternately in our method. In our alternating training strategy, the generator training iteration is unchanged. However, during the training iterations of $F$, we perform two sub-iterations (see Fig.~\ref{feedback_fig}). 

\noindent \textit{First sub-iteration}: The noise $z$ and semantic embeddings $a$ are input to the generator $G$ to yield an initial synthesized feature $\hat{x}[0]=G(z,a)$, which is then passed through to the semantic embedding decoder $Dec$. \\
\noindent \textit{Second sub-iteration}: The latent embedding $\hat{h}$ from $Dec$ is input to $F$, resulting in an output $\hat{x}^f[t] = F(\hat{h})$, which is added to the latent representation (denoted as $g^l$ in Eq.~\ref{eqn_feedback_concept}) of $G$. The same $z$ and $a$ (used in the first sub-iteration) are used as input to $G$ for the second sub-iteration, with the additional input $\hat{x}^f[t]$ added to the latent representation $g^l$ of generator $G$. The generator then outputs a synthesized feature $\hat{x}[t+1]$, as,
\begin{equation}
    \hat{x}[t+1] = G(z,a,\hat{x}^f[t]).
\end{equation}
The refined feature $\hat{x}[t+1]$ is input to $D$ and $Dec$, and corresponding losses are computed (Eq.~\ref{eqn_total_loss}) for training. In practice, the second sub-iteration is performed only once.
The feedback module $F$ allows generator $G$ to view the latent embedding of $Dec$, corresponding to current generated features. This enables $G$ to appropriately refine its output (feature generation) iteratively, leading to an enhanced feature representation.

\subsection{(Generalized) Zero-Shot Classification}
In our \proposed, unseen class features are synthesized by inputting respective embeddings $a(u)$ and noise $z$ to $G$, given by $\hat{x}_u=G(z,a(u),\hat{x}^f[0])$. Here, $\hat{x}^f[0]$ denotes feedback output of $F$, computed for the same $a(u)$ and $z$. The synthesized unseen class features $\hat{x}_u$ and real seen class features $x_s$ are further input to $Dec$ to obtain their respective latent embeddings, which are concatenated with input features. In this way, we obtain transformed features $x_s \oplus h_s$ and $\hat{x}_u \oplus \hat{h}_u$, which are used to train ZSL and GZSL classifiers, $f_{zsl}$ and $f_{gzsl}$, respectively. At inference, test features $x_t$ are transformed in a similar manner, to obtain $x_t \oplus h_t$. The transformed features are then input to classifiers for final predictions.

\section{Experiments}
\noindent\textbf{Datasets:} We evaluate our \proposed~framework on four standard zero-shot object recognition datasets: Caltech-UCSD-Birds~\cite{cub} (CUB), Oxford Flowers~\cite{flo} (FLO), SUN Attribute~\cite{sun} (SUN), and Animals with Attributes2~\cite{zsl-good-bad-ugly} (AWA2) containing $200$, $102$, $717$ and $50$ categories, respectively. For fair comparison, we use the \emph{same} splits, evaluation protocols and class embeddings as in~\cite{zsl-good-bad-ugly}.\\
\noindent\textbf{Visual features and embeddings:} We extract the average-pooled feature instances of size $2048$ from the ImageNet-1K~\cite{imagenet} pre-trained ResNet-101~\cite{resnet}. For semantic
 embeddings, we use the class-level attributes for CUB ($312$-d), SUN ($102$-d) and AWA2 ($85$-d). For FLO, fine-grained visual descriptions of image are used to extract $1024$-d embeddings from a character-based CNN-RNN~\cite{reed16cvpr}. \\
\noindent\textbf{Implementation details:} The discriminator $D$, encoder $E$ and generator $G$ are implemented as two-layer fully-connected (FC) networks with $4096$ hidden units. 
The dimensions of $z$ and $a$ are set to be equal ($\mathbb{R}^{d_z}=\mathbb{R}^{d_a}$). The semantic embedding decoder $Dec$ and feedback module $F$ are also two-layer FC networks with $4096$ hidden units. The input and output dimensions of $F$ are set to $4096$ to match the hidden units of $Dec$ and $G$. For transductive setting, an unconditional discriminator $D2$ is employed for utilizing the unlabelled feature instances during training, as in~\cite{Xian19cvpr}. Since the corresponding semantic embeddings are not available for unlabelled instances, only the visual feature is input to $D2$. Leaky ReLU activation is used everywhere, except at the output of $G$, where a \textit{sigmoid} activation is used for applying BCE loss. The network is trained using the Adam optimizer with $10^{-4}$ learning rate. Final ZSL/GZSL classifiers are single layer FC networks with output units equal to number of test classes. Hyper-parameters $\alpha$, $\beta$ and $\delta$ are set to $10$, $0.01$ and $1$, respectively. The gradient penalty coefficient $\lambda$ is initialized to $10$ and WGAN is trained, similar to~\cite{wgan}.

\subsection{State-of-the-art Comparison\label{sec_sota_compare}}
\begin{table}[t]
\centering
\caption{\label{tab_sota_compare}\textbf{State-of-the-art comparison on four datasets.} Both inductive (IN) and transductive (TR) results are shown. The results with fine-tuning the backbone network using the seen classes only (without violating ZSL), are reported under fine-tuned inductive (FT-IN) and transductive (FT-TR) settings. For ZSL, results are reported in terms of average \emph{top-1} classification accuracy (\textbf{T1}). For GZSL, results are reported in terms of \emph{top-1} accuracy of unseen ($u$) and seen ($s$) classes, together with their harmonic mean (\textbf{H}). Our \proposed~performs favorably in comparison to existing methods on \emph{all} four datasets, in all settings (IN, TR, FT-IN and FT-TR), for \emph{both} ZSL and GZSL.}%
\adjustbox{width=\textwidth}{
\begin{tabular}{llcccc|ccc|ccc|ccc|ccc}
 &  & \multicolumn{4}{c|}{\textbf{Zero-shot Learning}} & \multicolumn{12}{c}{\textbf{Generalized Zero-shot Learning}} \\
 &  & \multicolumn{1}{c}{\textbf{CUB}} & \multicolumn{1}{c}{\textbf{FLO}} & \multicolumn{1}{c}{\textbf{SUN}} & \multicolumn{1}{c|}{\textbf{AWA}} & \multicolumn{3}{c}{\textbf{CUB}} & \multicolumn{3}{c}{\textbf{FLO}} & \multicolumn{3}{c}{\textbf{SUN}} & \multicolumn{3}{c}{\textbf{AWA}} \\
 & & \multicolumn{1}{c}{\textbf{T1}} & \multicolumn{1}{c}{\textbf{T1}} & \multicolumn{1}{c}{\textbf{T1}} & \multicolumn{1}{c|}{\textbf{T1}} & \multicolumn{1}{c}{$u$} & \multicolumn{1}{c}{$s$} & \multicolumn{1}{c|}{\textbf{H}} & \multicolumn{1}{c}{$u$} & \multicolumn{1}{c}{$s$} & \multicolumn{1}{c|}{\textbf{H}} & \multicolumn{1}{c}{$u$} & \multicolumn{1}{c}{$s$} & \multicolumn{1}{c|}{\textbf{H}} & \multicolumn{1}{c}{$u$} & \multicolumn{1}{c}{$s$} & \multicolumn{1}{c}{\textbf{H}} \\ 
 \hline
\multirow{6}{*}{IN} & \clswgan~\cite{Xian18cvpr} & 57.3 & 67.2 & 60.8 & 68.2 & 3.7 & 57.7 & 49.7 & 59.0 & 73.8 & 65.6 & 42.6 & 36.6 & 39.4 & 57.9 & 61.4 &  59.6  \\
 & \texttt{Cycle-WGAN}~\cite{Rafael18eccv} & 58.6 & 70.3 & 59.9 & 66.8 & 47.9 & 59.3 & 53.0 & 61.6 & 69.2 & 65.2 & \textbf{47.2} & 33.8 & 39.4 & 59.6 & 63.4 & 59.8  \\
 & \texttt{LisGAN}~\cite{li19leveraging} & 58.8 & 69.6 & 61.7 & 70.6 & 46.5 & 57.9 & 51.6 & 57.7 & 83.8 & 68.3 & 42.9 & 37.8 & 40.2 & 52.6 & \textbf{76.3} & 62.3  \\
 & \texttt{TCN}~\cite{jiang2019transferable} & 59.5 & - & 61.5 & 71.2 & 52.6 & 52.0 & 52.3 & - & - & - & 31.2 & 37.3 & 34.0 & \textbf{61.2} & 65.8 &  63.4  \\
 & \vaegan~\cite{Xian19cvpr} & 61.0 & 67.7 & 64.7 & 71.1 & 48.4 & 60.1 & 53.6 & 56.8 & 74.9 & 64.6 & 45.1 & 38.0 & 41.3 & 57.6 & 70.6 &  63.5  \\
 & \texttt{Ours}: ${\proposed}$ & \textbf{64.9} & \textbf{70.8} & \textbf{66.0}  & \textbf{72.2}  & \textbf{52.8}  &  \textbf{64.7} & \textbf{58.1}  & \textbf{62.5}  & \textbf{84.1}  & \textbf{71.7} & 45.6  & \textbf{40.7}  & \textbf{43.0}  & 59.8 & 75.1 & \textbf{66.6}  \\
 \hline
\multirow{5}{*}{TR} & \texttt{ALE-tran}~\cite{zsl-good-bad-ugly} & 54.5 & 48.3 & 55.7 & 70.7 & 23.5 & 45.1 & 30.9 & 13.6 & 61.4 & 22.2 & 19.9 & 22.6  & 21.2 & 12.6 & 73.0 & 21.5 \\
 & \texttt{GFZSL}~\cite{verma17simple} & 50.0 & 85.4 & 64.0 & 78.6 & 24.9 & 45.8 & 32.2 & 21.8 & 75.0 & 33.8 & 0.0 & 41.6 & 0.0 & 31.7 & 67.2 & 43.1 \\
 & \texttt{DSRL}~\cite{Ye17cvpr} & 48.7 & 57.7 & 56.8 & 72.8 & 17.3 & 39.0 & 24.0 & 26.9 & 64.3 & 37.9 & 17.7 & 25.0 & 20.7 & 20.8 & 74.7 & 32.6  \\
 & \vaegan~\cite{Xian19cvpr} & 71.1 & 89.1 & 70.1 & 89.8 & 61.4 & 65.1 &  63.2 & 78.7 & 87.2 & 82.7 & 60.6 & 41.9 & 49.6 & 84.8 & 88.6  & 86.7  \\
 & \texttt{Ours}: ${\proposed}$ & \textbf{74.7} & \textbf{92.6} & \textbf{70.9}  & \textbf{92.1} & \textbf{69.9}  & \textbf{72.1} & \textbf{71.0}  & \textbf{91.8}  & \textbf{93.2} & \textbf{92.5}  & \textbf{62.4} & \textbf{47.1} & \textbf{53.7} & \textbf{87.3}  & \textbf{89.6}  & \textbf{88.4} \\
\hline \hline
\multirow{3}{*}{FT-IN} & \texttt{SBAR-I}~\cite{paul2019semantically} & 63.9 & - & 62.8 & 65.2 & 55.0 & 58.7 & 56.8 & - & - & - & \textbf{50.7} & 35.1 & 41.5 & 30.3 & \textbf{93.9} & 46.9  \\
& \texttt{\vaegan}~\cite{Xian19cvpr} & 72.9 &70.4& 65.6 &70.3& 63.2& 75.6& 68.9& 63.3 &92.4& 75.1 & 50.1& 37.8 & 43.1& \textbf{57.1}& 76.1& 65.2 \\
& \texttt{Ours}: ${\proposed}$ & \textbf{74.3} & \textbf{74.7} & \textbf{66.7}  & \textbf{73.4} & \textbf{63.8}  & \textbf{79.3} & \textbf{70.7} & \textbf{69.5} & \textbf{92.5}  & \textbf{79.4} & 41.8  & \textbf{51.9}  & \textbf{46.3}  & 55.5 & 83.6 & \textbf{66.7} \\
 \hline
\multirow{4}{*}{FT-TR} & 
\texttt{SBAR-T}~\cite{paul2019semantically} & 74.0 & - & 67.5 & 88.9 & 67.2 & 73.7 & 70.3 & - & - & - & \textbf{58.8} & 41.5 & 48.6 & 79.7 & \textbf{91.0} & 85.0  \\

& \texttt{UE-finetune}~\cite{song2018transductive} & 72.1 & - & 58.3 & 79.7 & 74.9 & 71.5 & 73.2 & - & - & -  &33.6 &54.8 &41.7& \textbf{93.1}& 66.2& 77.4 \\
& \texttt{\vaegan}~\cite{Xian19cvpr} &82.6 &95.4& 72.6& 89.3& 73.8& 81.4& 77.3& 91.0& 97.4& 94.1& 54.2& 41.8& 47.2& 86.3& 88.7& 87.5 \\
& \texttt{Ours}: ${\proposed}$ & \textbf{85.1} &\textbf{96.0} & \textbf{73.8}  & \textbf{93.0}  & \textbf{78.4}  & \textbf{83.5} & \textbf{80.9} & \textbf{96.1}  & \textbf{97.6} & \textbf{96.8} & 44.3  & \textbf{66.9} &  \textbf{53.3} & 89.2 & 90.0 &  \textbf{89.6} \\
\end{tabular}
}
\end{table}%

Tab.~\ref{tab_sota_compare} shows state-of-the-art comparison on four object recognition datasets. Results for inductive (IN) and transductive (TR) settings are obtained without any fine-tuning of the backbone network. For inductive (IN) ZSL, the \texttt{Cycle-WGAN}~\cite{Rafael18eccv} obtains classification scores of $58.6\%$, $70.3\%$, $59.9\%$, and $66.8\%$ on CUB, FLO, SUN and AWA, respectively. The \vaegan~\cite{Xian19cvpr} reports classification accuracies of $61\%$, $67.7\%$, $64.7\%$, and $71.1\%$ on the same datasets. Our \proposed{} outperforms \vaegan{} on \emph{all} datasets achieving classification scores of $64.9\%$, $70.8\%$, $66.0\%$, and $72.2\%$ on CUB, FLO, SUN and AWA, respectively. 
In the transductive (TR) ZSL setting, \vaegan{} obtains \emph{top-1} classification (\textbf{T1}) accuracies of $71.1\%$, $89.1\%$, $70.1\%$, and $89.8\%$ on the four datasets. Our \proposed{} outperforms \vaegan{} on \emph{all} datasets, achieving classification accuracies of $74.7\%$, $92.6\%$, $70.9\%$, and $92.1\%$ on CUB, FLO, SUN and AWA, respectively. Similarly, our \proposed{} also performs favourably compared to existing methods on all datasets for both inductive and transductive GZSL settings. Utilizing unlabelled instances during training, to reduce the domain shift problem for unseen classes, in the transductive setting yields higher results compared to inductive setting.

Some previous works, including \vaegan~\cite{Xian19cvpr} have reported results with fine-tuning the backbone network only using the seen classes (without violating the ZSL condition). Similarly, we also evaluate our \proposed{} by utilizing fine-tuned backbone features. Tab.~\ref{tab_sota_compare} shows the comparison with existing fine-tuning based methods for both ZSL and GZSL in fine-tuned inductive (FT-IN) and fine-tuned transductive (FT-TR) settings. For FT-IN ZSL, \vaegan~obtains classification scores of $72.9\%$, $70.4\%$, $65.6\%$, and $70.3\%$ on CUB, FLO, SUN and AWA, respectively. Our \proposed{} achieves consistent improvement over \vaegan{} on \emph{all} datasets, achieving classification scores of $74.3\%$, $74.7\%$, $66.7\%$, and $73.4\%$ on CUB, FLO, SUN and AWA, respectively. Our approach also improves over \vaegan{} for the FT-TR ZSL setting. In the case of FT-IN GZSL, our \proposed{} achieves gains (in terms of \textbf{H}) of $1.8\%$, $4.3\%$, $3.2\%$, and $1.5\%$ on CUB, FLO, SUN and AWA, respectively over \vaegan. A similar trend is also observed for the FT-TR GZSL setting. 
In summary, our \proposed{} achieves promising results for various settings and backbone feature combinations.

\begin{table*}[t]
\centering
\caption{\label{tab_baseline_compare}\textbf{Baseline performance comparison} on CUB~\cite{cub}. In both inductive and transductive settings, our \feedback~and \featcat~provide consistent improvements over the baseline for both ZSL and GZSL. Further, our final \proposed~framework, integrating both \feedback~and \featcat, achieves further gains over the baseline in both inductive and transductive settings, for ZSL and GZSL.}
\adjustbox{width=\textwidth}{
\begin{tabular}{ccccc|cccc}

 & \multicolumn{4}{c|}{\texttt{INDUCTIVE}} & \multicolumn{4}{c}{\texttt{TRANSDUCTIVE}} \\ 
\multicolumn{1}{l|}{} & \multicolumn{1}{l}{\texttt{Baseline}} & \multicolumn{1}{l}{\feedback~} & \multicolumn{1}{l}{\featcat~} & \multicolumn{1}{l|}{\proposed~} & \multicolumn{1}{l}{\texttt{Baseline}} & \multicolumn{1}{l}{\feedback~} & \multicolumn{1}{l}{\featcat~} & \multicolumn{1}{l}{\proposed~} \\ \hline
\multicolumn{1}{c|}{\texttt{\textbf{ZSL}}} & 61.2 & 62.8 & 64.0 & \textbf{64.9} & 70.6 & 71.7 & 73.5 & \textbf{74.7} \\ 
\multicolumn{1}{c|}{\texttt{\textbf{GZSL}}} & 53.5 & 54.8 & 56.9 & \textbf{58.1} & 63.7 & 66.8 & 69.2 & \textbf{71.0} \\ \hline
\end{tabular}
}
\end{table*}

\subsection{Ablation Study\label{sec_ablation}}
\begin{figure}[t]
\centering
\includegraphics[width=0.79\columnwidth]{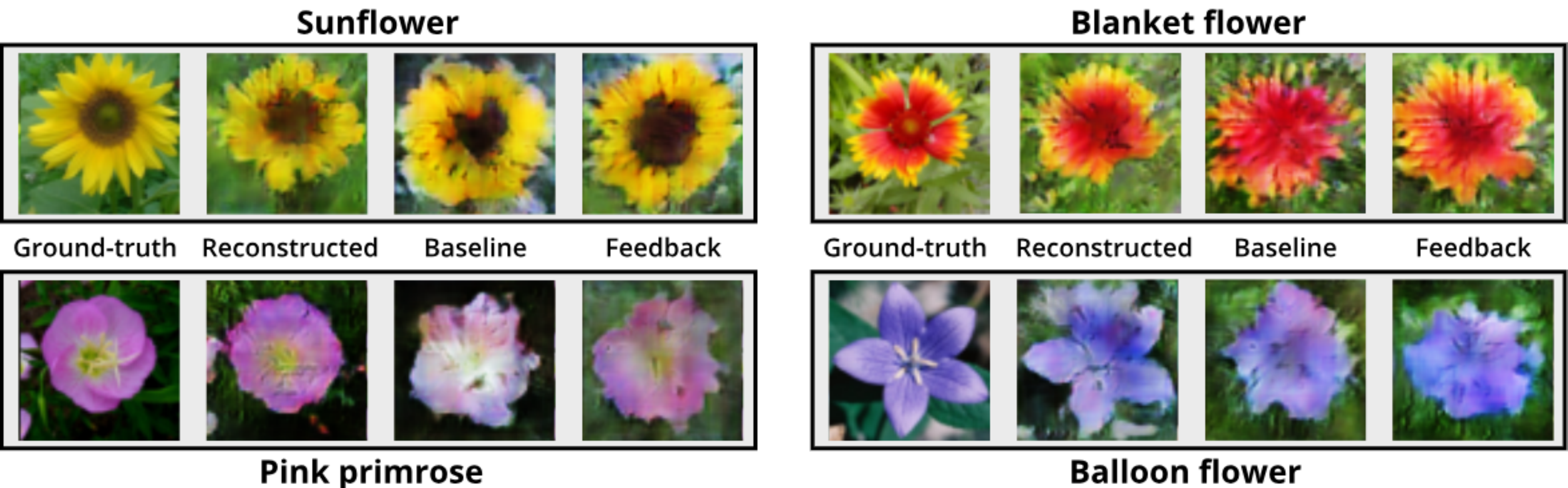}
\caption{\label{fig:qual_res}\textbf{Qualitative comparison} between inverted images of \texttt{Baseline} synthesized features and our \texttt{Feedback} synthesized features on four example classes of FLO~\cite{flo}. The ground-truth image and the reconstructed inversion of its real feature are also shown for each example. Our \texttt{Feedback} improves petal shapes (\textit{Sunflower}), shape of bud and petals (\textit{Blanket flower}), color (\textit{Pink primrose}), black lining on petals (\textit{Balloon flower}) and achieves promising improvements over \texttt{Baseline}. Best viewed in zoom.}%
\end{figure}

\noindent\textbf{Baseline comparison}: We first compare our proposed \proposed{} with the baseline \vaegan~\cite{Xian19cvpr} on CUB for (generalized) zero-shot recognition in both inductive and transductive settings. The results are reported in Tab.~\ref{tab_baseline_compare} in terms of average \emph{top-1} classification accuracy for ZSL and harmonic mean of the classification accuracies of seen and unseen classes for GZSL. For the baseline, we present the results based on our re-implementation. In addition to our final \proposed, we report results of our feedback module alone (denoted 
\begin{wrapfigure}{r}{0.5\columnwidth}
\centering
\includegraphics[width=0.5\columnwidth, trim={0 0.1cm 0 0.11cm},clip]{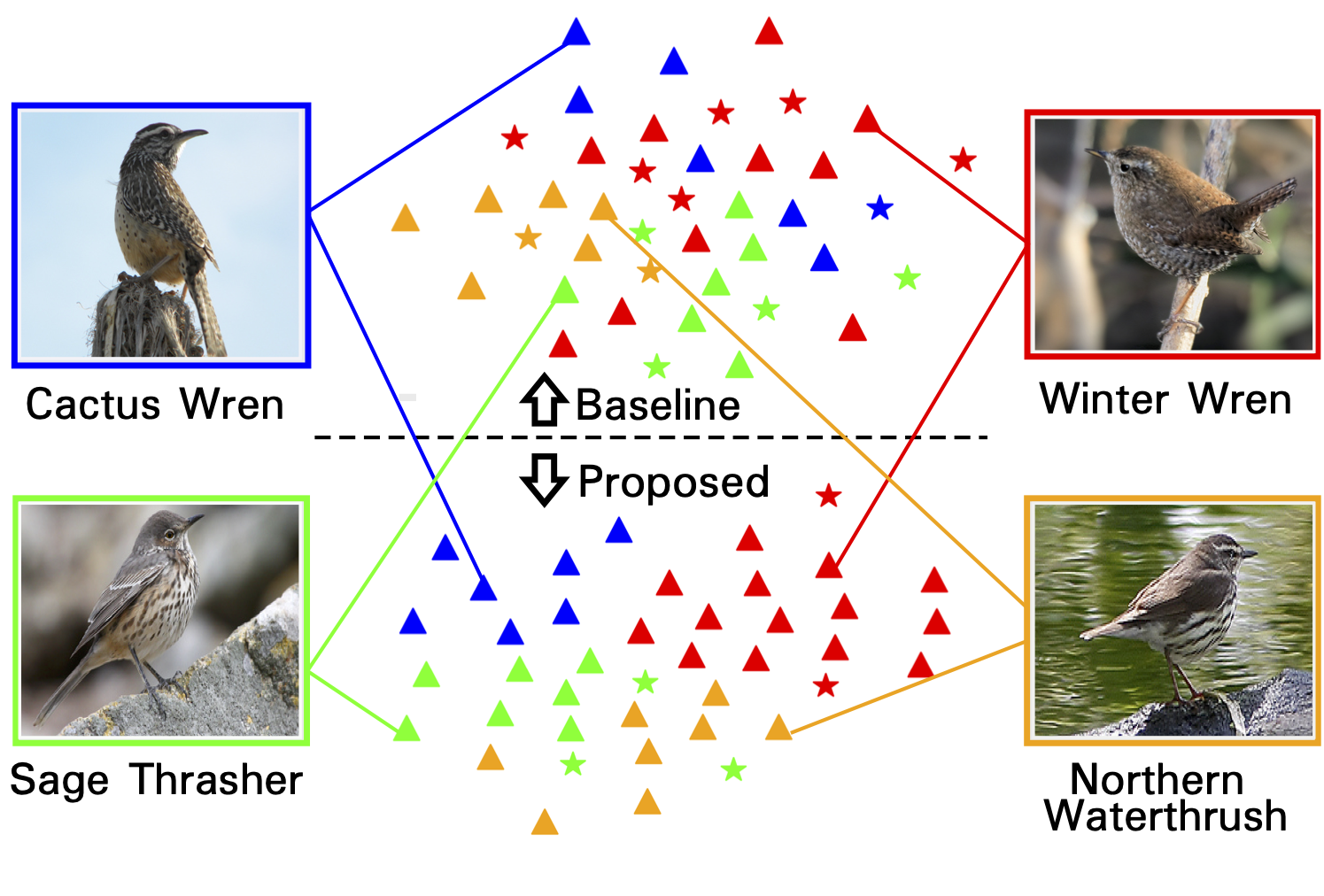}%
\caption{\label{fig_qual}\textbf{t-SNE visualization} of test instances of four fine-grained classes in CUB~\cite{cub} dataset. Both \texttt{Cactus Wren} and \texttt{Winter Wren} belong to the same family \texttt{Troglodytidae}. Further, \texttt{Cactus Wren} is visually similar to \texttt{Sage Thrasher} and \texttt{Northern Waterthrush}. Top: the baseline method struggles to correctly classify instances of these categories (denoted by $\bigstar$ with respective class color) due to inter-class confusion. Bottom: our approach improves the inter-class grouping and decreases misclassifications, leading to favourable performance.}%
\end{wrapfigure}
as \feedback~in Tab.~\ref{tab_baseline_compare}) without feature transformation utilized at classification stage. Moreover, the performance of discriminative feature transformation alone (denoted as \featcat), without utilizing the feedback is also presented. 
For the inductive setting, \texttt{Baseline} obtains a classification performance of $61.2\%$ and $53.5\%$ for ZSL and GZSL. Both our contributions, \feedback{} and \featcat{}, consistently improve the performance over the baseline. The best results are obtained by our \proposed, with gains of $3.7\%$ and $4.6\%$ over the baseline, for ZSL and GZSL. Similar to the inductive (IN) setting, our proposed \proposed~also achieves favourable performance in transductive (TR) setting. Fig.~\ref{fig_qual} shows a comparison between baseline and our \proposed~methods, using t-SNE visualizations~\cite{maaten08tsne} of test instances from four example fine-grained classes of CUB. While the baseline struggles to correctly classify these fine-grained class instances due to inter-class confusion, our \proposed~improves inter-class grouping leading to a favorable classification performance.\\
\begin{figure}[t]
\begin{minipage}{0.5\textwidth}
    \centering
    \subfloat[ZSL]%
    {\includegraphics[width=0.72\textwidth]{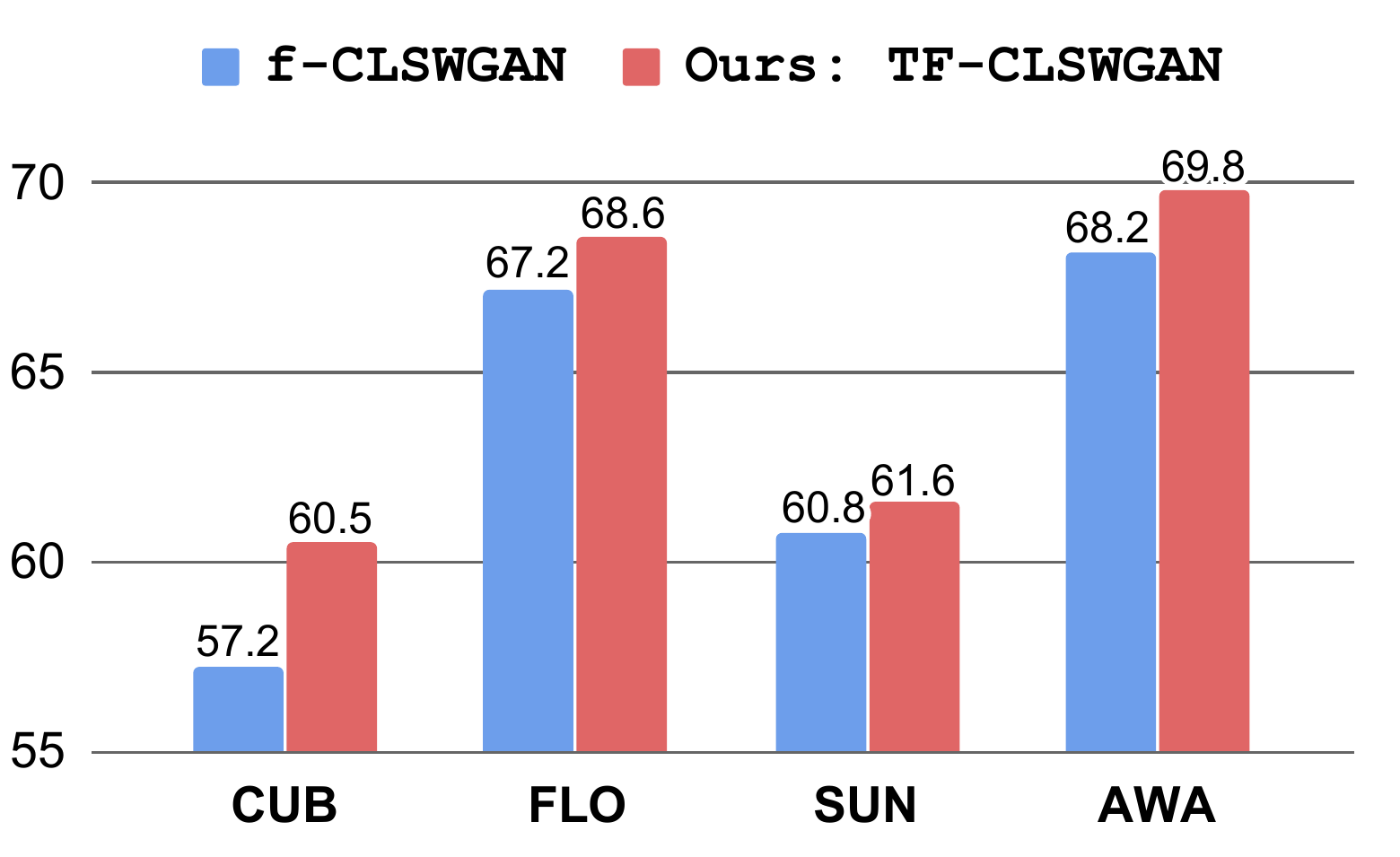}%
    }
\end{minipage}%
\begin{minipage}{0.5\textwidth}
    \centering
    \subfloat[GZSL]%
    {\includegraphics[width=0.72\textwidth]{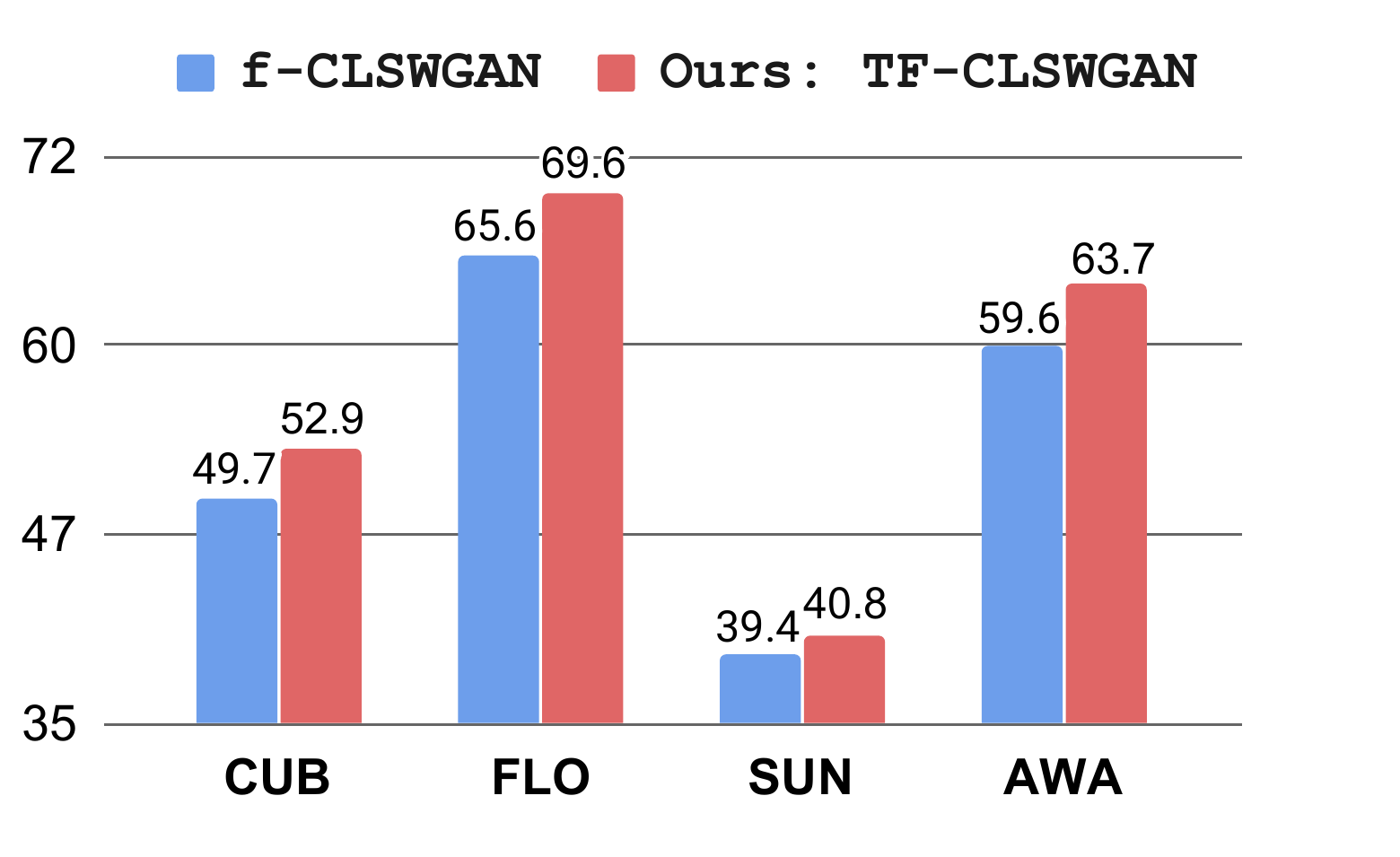}%
    }
\end{minipage}%
\caption{\label{generalization_overall} \textbf{Generalization capabilities.} (a) ZSL and (b) GZSL performance comparison to validate the generalization capabilities of our contributions. Instead of a VAE-GAN architecture, we integrate our proposed contributions in the \clswgan~framework. Our \texttt{TF-CLSWGAN} outperforms the vanilla \clswgan~on all datasets. Best viewed in zoom.}%
\end{figure}%
\noindent\textbf{Generalization capabilities}: Here, we base our approach on a VAE-GAN architecture~\cite{Xian19cvpr}. However, our proposed contributions (a semantic embedding decoder at all stages of the ZSL pipeline and the feedback module) are generic and can also be utilized in other GAN-based ZSL frameworks. To this end, we perform an experiment by integrating our contributions in the \clswgan~\cite{Xian18cvpr} architecture. Fig.~\ref{generalization_overall} shows the comparison between the baseline \clswgan~and our \texttt{TF-CLSWGAN} for ZSL and GZSL tasks, on all four datasets. Our \texttt{TF-CLSWGAN} outperforms the vanilla \clswgan~in all cases for both ZSL and GZSL tasks. \\
\noindent\textbf{Feature visualization:} To qualitatively assess the feature synthesis stage, we train an upconvolutional network to invert the feature instances back to the image space by following a similar strategy as in~\cite{dosovitskiy2016generating,Xian19cvpr}. 
Corresponding implementation details are provided in the supplementary. The model is trained on all real feature-image pairs of the 102 classes of FLO~\cite{flo}. The comparison between \texttt{Baseline} and our \texttt{Feedback} synthesized features on four example flowers is shown in Fig.~\ref{fig:qual_res}. For each flower class, a ground-truth (GT) image along with three images inverted from its GT feature, \texttt{Baseline} and \texttt{Feedback} synthesized features, respectively are shown. Generally, inverting the \texttt{Feedback} synthesized feature yields an image that is semantically closer to the GT image than inverting the \texttt{Baseline} synthesized feature. This suggests that our \texttt{Feedback} improves the feature synthesis stage over the \texttt{Baseline}, where no feedback is present. \\ 
Additional quantitative and qualitative results are given in the supplementary.

\section{(Generalized) Zero-Shot Action Recognition}
Finally, we validate our \proposed~for action recognition in videos under ZSL and GZSL.
Here, we use the I3D features~\cite{carreira17i3d}, as in the GAN-based zero-shot action classification method \texttt{CEWGAN}~\cite{Mandal19cvpr}. While using improved video features is likely to improve the performance of a zero-shot action recognition framework, our goal is to show that our \proposed{} generalizes to action classification and improves the performance using the same underlying video features. As in~\cite{Mandal19cvpr}, we extract spatio-temporally pooled 4096-d I3D features from pre-trained RGB and Flow I3D networks and concatenate them to obtain 8192-d video features. Further, an out-of-distribution classifier is utilized at the classification stage, as in~\cite{Mandal19cvpr}. For HMDB51, a skip-gram model~\cite{mikolov13w2v} is used to generate semantic embeddings of size $300$, using action class names as input. For UCF101, we use semantic embeddings of size $115$, provided with the dataset. \\
Tab.~\ref{tab_action_sota_compare} shows state-of-the-art comparison on HMDB51~\cite{hmdb51} and UCF101~\cite{ucf101}. For a fair comparison, we use the same splits, embeddings and evaluation protocols as in~\cite{Mandal19cvpr}. On HMDB51, \vaegan~obtains classification scores of $31.1\%$ and $35.6\%$ for ZSL and GZSL. The work of~\cite{uar18cvpr} provides classification results of $24.4\%$ and $17.5\%$ for HMDB51 and UCF101, respectively for ZSL. Note that~\cite{uar18cvpr} also reports results using cross-dataset training on large-scale ActivityNet~\cite{activitynet}. On HMDB51, \texttt{CEWGAN}~\cite{Mandal19cvpr} obtains $30.2\%$ and $36.1\%$ for ZSL and GZSL. Our \proposed~achieves $33.0\%$ and $37.6\%$ for ZSL and GZSL. Similarly, our approach~performs favourably compared to existing methods on UCF101. Hence, our \proposed{} generalizes to action recognition and achieves promising results. 

\begin{table*}[t]
\centering
\caption{\label{tab_action_sota_compare}\textbf{State-of-the-art ZSL and GZSL comparison for action recognition}. Our \proposed~performs favorably against all existing methods, on both datasets.}
\adjustbox{width=\textwidth}{
\begin{tabular}{cc|c|c|c|c|c|c|c}
\multicolumn{1}{l}{} & \multicolumn{1}{l|}{} & \multicolumn{1}{l|}{\texttt{GGM}~\cite{Mishra18wacv}} & \multicolumn{1}{l|}{\texttt{CLSWGAN}~\cite{Xian18cvpr}} & \multicolumn{1}{l|}{\texttt{CEWGAN}~\cite{Mandal19cvpr}} & \multicolumn{1}{l|}{\texttt{Obj2Act}~\cite{jain15iccv}} & \multicolumn{1}{l|}{\texttt{ObjEmb}~\cite{mettes17iccv}} & \multicolumn{1}{l|}{\vaegan~\cite{Xian19cvpr}} & \multicolumn{1}{l}{\proposed~} \\ \hline
\multirow{2}{*}{HMDB51} & \texttt{\textbf{ZSL}} & 20.7 & 29.1 & 30.2 & 24.5 & - & 31.1 & \textbf{33.0} \\ 
 & \texttt{\textbf{GZSL}} & 20.1 & 32.7 & 36.1 & - & - & 35.6 & \textbf{37.6} \\ \hline
\multirow{2}{*}{UCF101} & \texttt{\textbf{ZSL}} & 20.3 & 37.5 & 38.3 & 38.9 & 40.4 & 38.2 & \textbf{41.0} \\ 
 & \texttt{\textbf{GZSL}} & 17.5 & 44.4 & 49.4 & - & - & 47.2 & \textbf{50.9} \\ 
\end{tabular}%
}
\end{table*}%

\section{Conclusion}
We propose an approach that utilizes the semantic embedding decoder (SED) at all stages (training, feature synthesis and classification) of a VAE-GAN based ZSL framework. Since SED performs inverse transformations in relation to the generator, its deployment at all stages enables exploiting complementary information with respect to feature instances. To effectively utilize SED during both training and feature synthesis, we introduce a feedback module that transforms the latent embeddings of the SED and modulates the latent representations of the generator. We further introduce a discriminative feature transformation, during the classification stage, which utilizes the latent embeddings of SED along with respective features. Experiments on six datasets clearly suggest that our approach achieves favorable performance, compared to existing methods.

{\small
\bibliographystyle{ieee_fullname}
\bibliography{egbib}
}

\clearpage

\appendix
\section{Quantitative Results\label{sec_quant_res}}
In this section, we present the ablation studies with respect to the feedback design choices and the choice of latent embeddings.

\noindent\textbf{Feedback design choices}:
Here, we explore the effect of changing the input to the feedback module $F$ and its associated training strategy on CUB. Originally, the input to $F$ is taken from discriminator $D$ and the training of $F$ is performed in a two-stage strategy. This setup is denoted by \texttt{TwoStage+D} and obtains classification performance of $61.4\%$ and $53.3\%$ for ZSL and GZSL. Instead, in our approach, the input to $F$ is taken from SED $Dec$. This setup is denoted by \texttt{TwoStage+Dec} and achieves performance of $62.0\%$ and $53.8\%$ for ZSL and GZSL. Further, we utilize an alternate training strategy combined with \texttt{TwoStage+Dec} to facilitate the generator training, thereby improving feature synthesis. This setup, denoted by \texttt{Our Feedback}, achieves improved performance of $62.8\%$ and $54.8\%$ for ZSL and GZSL. These results show that (i) \texttt{TwoStage+Dec} provides improved performance over original \texttt{TwoStage+D} and (ii) the best results are obtained by \texttt{Our Feedback}, demonstrating the impact of our modifications for improved zero-shot recognition.

\noindent\textbf{Choice of latent embeddings for \featcat}: Here, we evaluate the impact of concatenating different embeddings from SED to the baseline features.  We compare our proposed concatenation (\featcat) of baseline features with latent embeddings $h$ of SED with both the original baseline features (\texttt{OrigFeat}) and the baseline features concatenated with the reconstructed attributes (\texttt{ConcatFeat}). On CUB, \texttt{OrigFeat} achieves $61.2\%$ and $53.5\%$ on ZSL and GZSL tasks, respectively. \texttt{ConcatFeat} achieves gains of $1.6\%$ and $2.0\%$ over \texttt{OrigFeat}.
In case of \texttt{ConcatFeat}, the reconstructed attributes have single feature representations per-class with inter-class separability but no intra-class diversity. Different to reconstructed attributes, the latent embeddings $h$ possess both intra-class diversity (multiple feature instances per class) and inter-class separability. Our \featcat~exploits these properties of latent embeddings with improved results over both \texttt{OrigFeat} and \texttt{ConcatFeat}. Compared to \texttt{OrigFeat}, \featcat~obtains gains of $2.8\%$ and $3.4\%$ on ZSL and GZSL tasks, respectively.

\section{Qualitative Analysis\label{sec_qual_res}}

\subsection{Feature Visualization Comparison\label{sec_feat_viz}}
Here, we present the implementation details and additional qualitative results for the visualization of synthesized features discussed in Sec.~4.2 of the paper. 

\noindent\textbf{Implementation details:} The image generator, which inverts the feature instances to images of size 64x64, consists of a fully-connected (FC) layer followed by five upconvolutional blocks. Each upconvolutional block contains an Upsampling layer, a 3x3 convolution, BatchNorm and ReLu non-linearity. An $\ell_1$ loss between the ground truth and inverted images, along with a perceptual loss ($\ell_2$ loss between the corresponding feature vectors at conv5 of a pre-trained ResNet-101) and an adversarial loss are employed to construct good quality images. The discriminator, required for adversarial training, takes image and feature embedding as inputs. The input image is processed through four downsampling blocks to obtain an image embedding, while the feature embedding is passed through an FC layer and spatially replicated to match the spatial dimensions of the obtained image embedding. The resulting two embeddings are concatenated and passed through convolutional and \textit{sigmoid} layers for predicting whether the input image is real or fake. The model is trained on all the real feature-image pairs of the 102 classes of FLO~\cite{flo}.

\begin{figure}[t]
\centering
\includegraphics[width=0.75\textwidth]{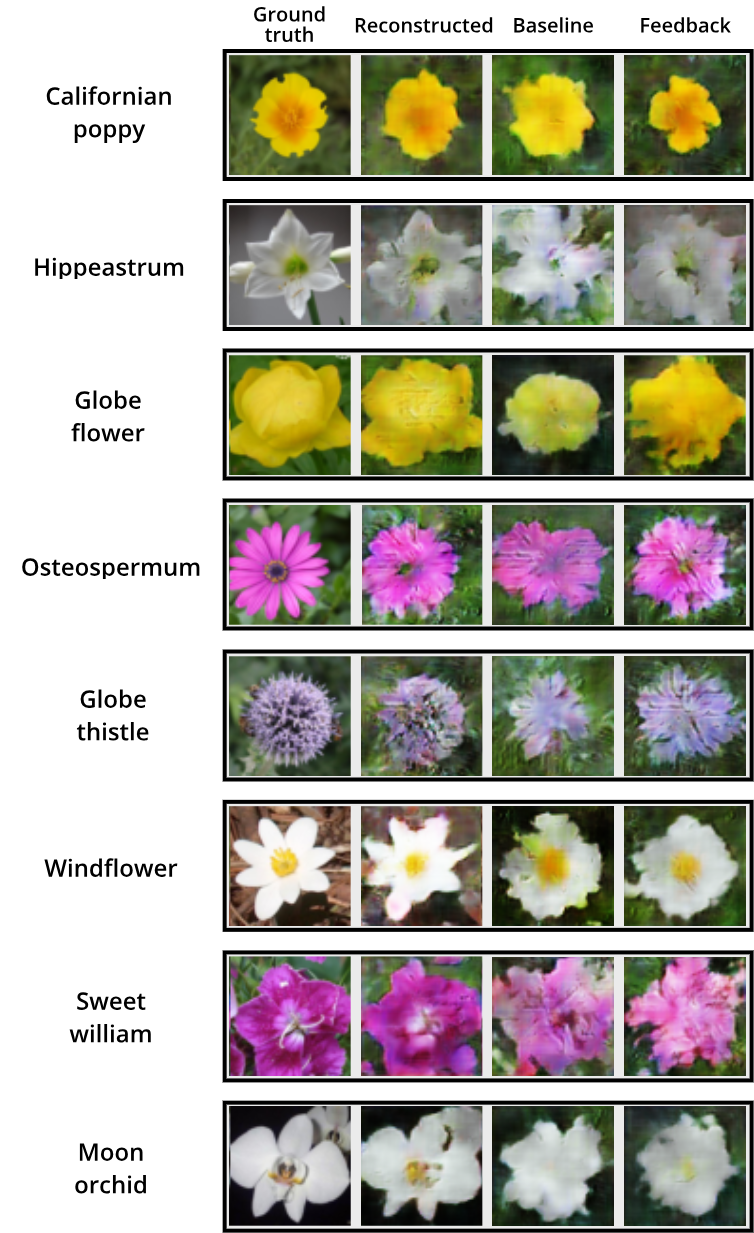}
\caption{\label{fig_feat_viz}\textbf{Qualitative comparison} between inverted images of \texttt{Baseline} synthesized features and our \texttt{Feedback} synthesized features on eight example classes of FLO~\cite{flo}. The ground-truth image (GT) and the reconstructed inversion (\texttt{Reconstructed}) of its real feature are also shown for each example. Inverting the feature instances from our \texttt{Feedback} improves the color of bud and shape of petals (\textit{Californian poppy}, \textit{Globe flower} and \textit{Osteospermum}), structure of the flower (\textit{Hippeastrum}), in comparison to the \texttt{Baseline} synthesized features. Semantic gap still exists between the inversion of real features (denoted as \texttt{Reconstructed}) and inversion of \feedback{} synthesized features for a few challenging cases (\eg, \textit{Globe thistle}, \textit{Windflower}, \textit{Sweet william}, \textit{Moon orchid}), even though there is some improvement for our \feedback{} over the \texttt{Baseline}. These observations suggest that our \texttt{Feedback} improves the quality of synthesized features over the \texttt{Baseline}, where no feedback is present. Best viewed in color and zoom.}
\end{figure}

\noindent\textbf{Visualization:} The comparison between \texttt{Baseline} and our \texttt{Feedback} synthesized features on eight example flowers is shown in Fig.~\ref{fig_feat_viz}. For each flower class, a ground-truth (GT) image along with three images inverted from its GT feature, \texttt{Baseline} and \texttt{Feedback} synthesized features, respectively are shown. Generally, inverting the \texttt{Feedback} synthesized feature yields an image that is semantically closer to the GT image than inverting the \texttt{Baseline} synthesized feature. Inverting the feature instances from our \texttt{Feedback} improves the color of bud and shape of petals (\textit{Californian poppy}, \textit{Globe flower} and \textit{Osteospermum}), structure of the flower (\textit{Hippeastrum}), in comparison to the \texttt{Baseline} synthesized features. A considerable improvement for our \texttt{Feedback} over the \texttt{Baseline} is visible in these flowers (\textit{Californian poppy}, \textit{Globe flower}, \textit{Hippeastrum} and \textit{Osteospermum}). However, there are a few challenging cases (\eg, \textit{Globe thistle}, \textit{Windflower}, \textit{Sweet william}, \textit{Moon orchid}), where a semantic gap still exists between the inversion of real features (denoted as \texttt{Reconstructed}) and inversion of \feedback{} synthesized features, even though there is a marginal improvement for our \feedback{} over the \texttt{Baseline}. These qualitative observations suggest that our \texttt{Feedback} improves the feature synthesis stage over the \texttt{Baseline}, where no feedback is present, resulting in improved zero-shot classification.

%

\clearpage

\subsection{Classification Performance Comparison\label{sec_cls_perf}}
\begin{figure}[t]
\centering
\includegraphics[width=\textwidth]{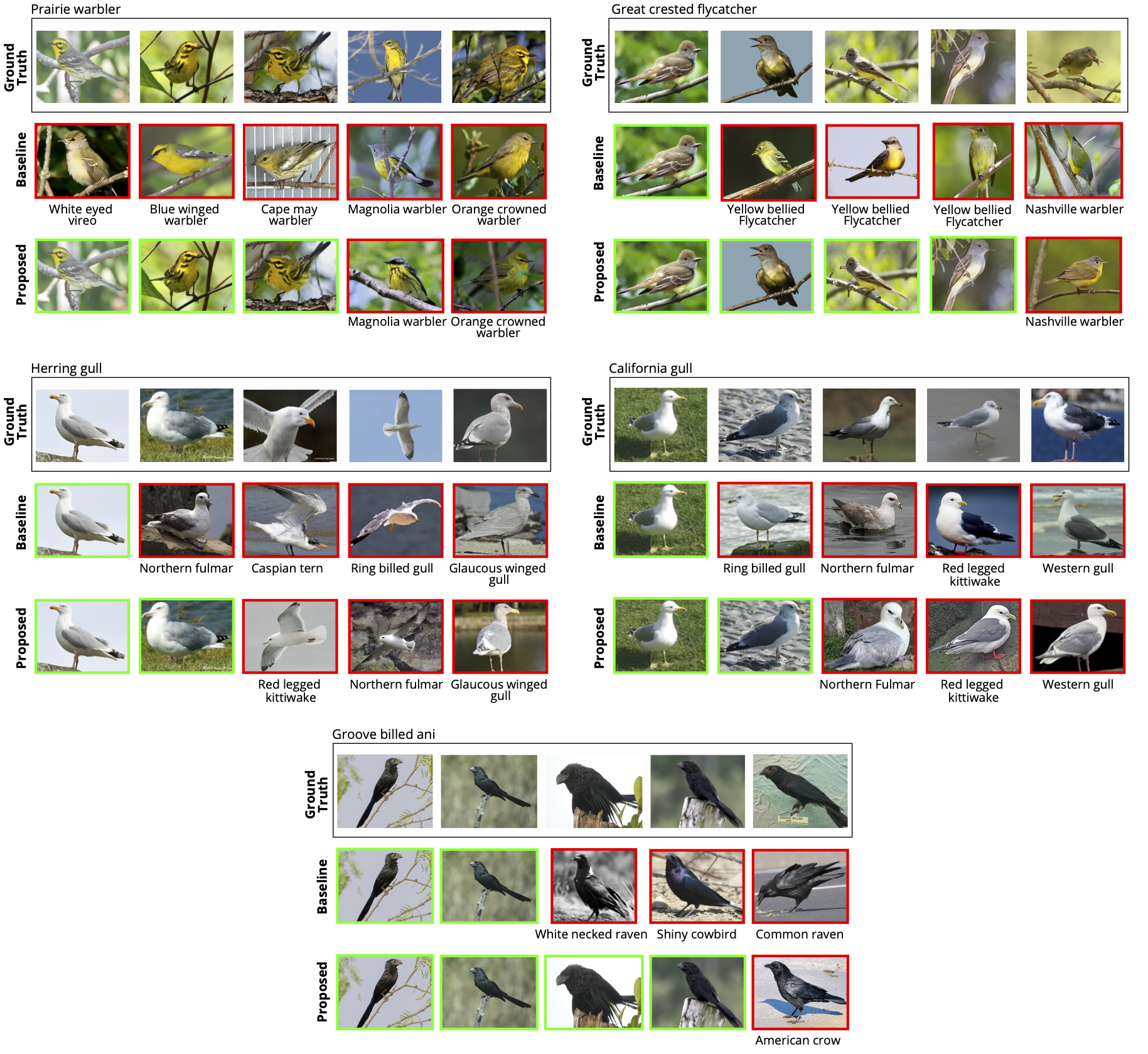}
\caption{\label{fig_cub_supp}Qualitative comparison between the baseline and our proposed approach on the CUB~\cite{cub} dataset. The comparison is based on the most confusing categories as per the baseline performance. For each category, while the top row denotes different variations of ground truth class instances, the second and third rows show the classification predictions by the baseline and proposed approaches, respectively. The green and red boxes denote correct and incorrect classification predictions, respectively. The class names under each red box show the corresponding incorrectly predicted label. In general, we observe that the instances are misclassified to other similar looking categories in the dataset. For instance, \textit{Prairie warbler} is confused with \textit{Blue winged warbler}, while \textit{Groove billed ani} is confused commonly with \textit{Common raven}.  For all these categories, the proposed \proposed~reduces the confusion among similar looking classes in the baseline \vaegan~and improves the classification performance over the baseline. See associated text for additional details. Best viewed in color and zoom.}
\end{figure}

\begin{figure}[t]
\centering
\includegraphics[width=\columnwidth]{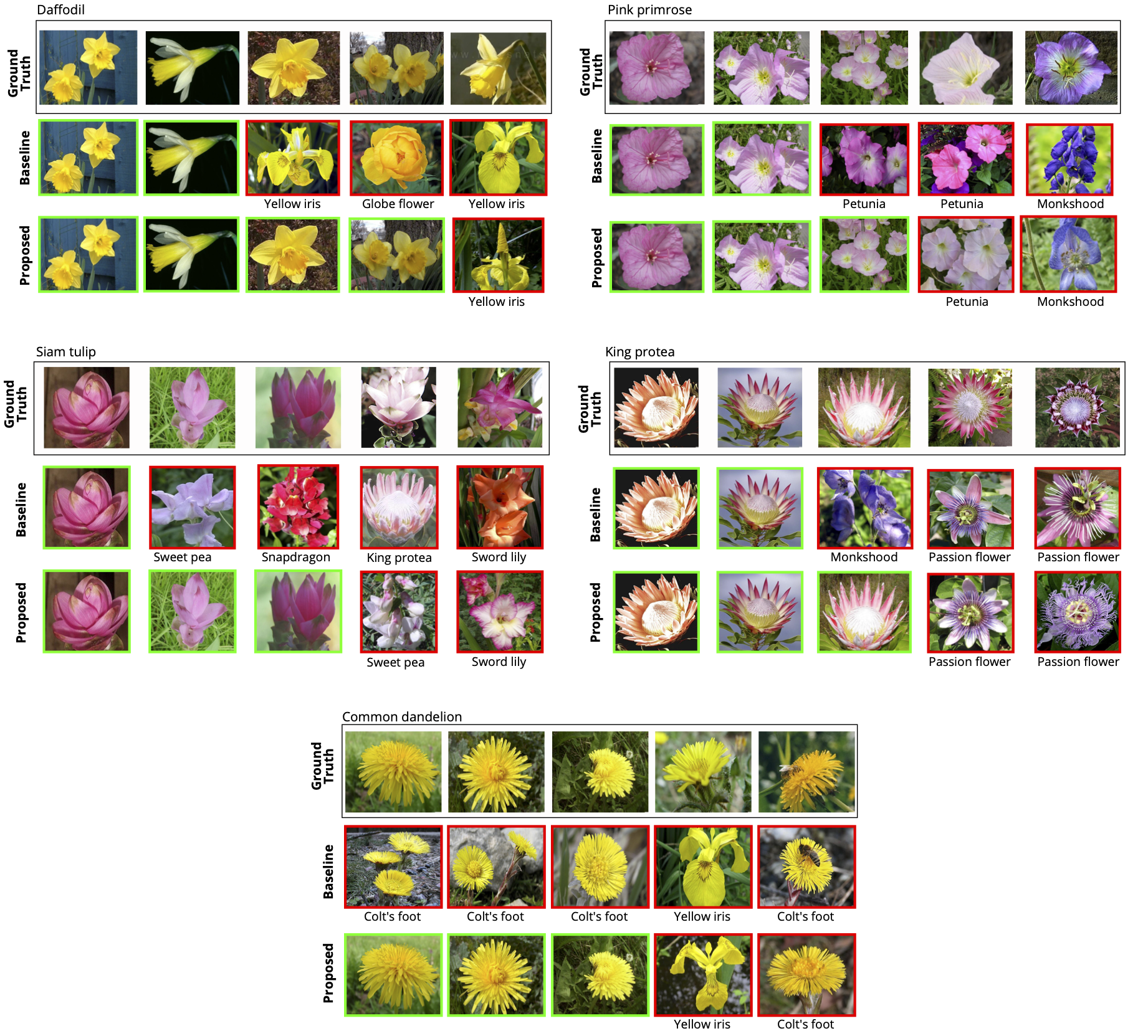}
\caption{\label{fig_flo_supp}Qualitative comparison between the baseline and our proposed approach on the Oxford Flowers~\cite{flo} dataset. The comparison is based on the most confusing categories as per the baseline performance. For each category, while the top row denotes different variations of ground truth class instances, the second and third rows show the classification predictions by the baseline and proposed approaches, respectively. The green and red boxes denote correct and incorrect classification predictions, respectively. The class names under each red box show the corresponding incorrectly predicted label. In general, we observe that the instances are misclassified to other similar looking categories in the dataset. For instance, \textit{Common dandelion} is confused with \textit{Colt's foot}, while \textit{Pink primrose} is confused with \textit{Petunia}. For all these categories, the proposed \proposed~reduces the confusion among similar looking classes in the baseline \vaegan~and improves the classification performance over the baseline. See associated text for additional details. Best viewed in color and zoom.}
\end{figure}

Here, we qualitatively illustrate the performance of our \proposed~framework, in comparison to the baseline \vaegan~\cite{Xian19cvpr} method, on two fine-grained object recognition datasets: CUB and FLO. Fig.~\ref{fig_cub_supp} and~\ref{fig_flo_supp} present the comparison on CUB and FLO, respectively. For each dataset, images from five most confusing categories (with respect to the baseline \vaegan) are shown. The comparison is illustrated for five image instances in each category. The ground truth instances are shown in the top row for each category, followed by the classification results of the baseline and proposed frameworks in second and third rows, respectively. Correctly classified images are marked with a green border, while the incorrectly classified images are marked with a red border. For the misclassifications, the name of the incorrectly predicted class is denoted below the instance for the respective methods. 

\noindent\textbf{CUB}: The qualitative comparison between the baseline and the proposed approaches for the CUB~\cite{cub} dataset is shown in Fig.~\ref{fig_cub_supp}. Five categories of birds that are most confusing for the baseline approach are presented. The categories are \textit{Prairie warbler}, \textit{Great crested flycatcher}, \textit{Grovve billed ani}, \textit{Herring gull} and \textit{California gull}. Generally, for all these categories, the baseline \vaegan~approach confuses with similar looking bird categories in the dataset. Our \proposed~reduces this confusion between similar looking classes and improves the classification performance. In Fig.~\ref{fig_cub_supp}, we observe that the baseline approach confuses \textit{Prairie warbler} class with other similar looking \textit{warbler} categories such as \textit{Blue winged warbler}, \textit{Magnolia warbler} and \textit{Orange crowned warbler}. This confusion is reduced in the predictions of our \proposed. Similarly, the confusion present, in the baseline method, between the \textit{Great crested flycatcher} and other \textit{flycatcher} categories is reduced for the proposed method. As a result, the overall classification performance improves for the proposed method over the baseline. 

\noindent\textbf{FLO}: Fig.~\ref{fig_flo_supp} shows the qualitative comparison for five categories of flowers from the Oxford Flowers~\cite{flo} dataset that are most confusing for the baseline method. The categories are \textit{Dafodil}, \textit{Pink primrose}, \textit{Siam tulip}, \textit{King Protea} and \textit{Common dandelion}. For all these categories, the proposed \proposed{} reduces the confusion present between the similar looking classes in the baseline \vaegan{} approach and improves the classification performance. In general, we observe that the instances are misclassified to other similar looking categories in the dataset. \Eg, instances of \textit{Common dandelion} are commonly misclassified as either \textit{Colt's foot} or \textit{Yellow iris}. All three categories have yellow flowers and share similar appearance. We observe that the baseline makes confused predictions with respect to these classes. However, the confusion is less in the predictions of the proposed \proposed. This leads to a favourable improvement in the zero-shot classification performance for the proposed approach. Similar observations can also be made in the case of other categories. The baseline \vaegan{} generally confuses \textit{Dafodil} with \textit{Globe flower} and \textit{Yellow iris} due to the yellow colour, while \textit{Pink primrose} is mostly confused with \textit{Petunia} and \textit{Monkshood} due to the pinkish petals in the flowers. The misclassifications are reduced when using the proposed \proposed{} for classification, resulting in an improved performance.

\clearpage

\clearpage

\end{document}